%% file: iclr2021_conference.tex
\pdfoutput=1
\documentclass{article} % For LaTeX2e
\usepackage{iclr2021_conference,times}

\input{math_commands.tex}

\usepackage{booktabs}    %导言区	

\usepackage{hyperref}
\usepackage{url}
\usepackage{graphicx} 
\usepackage{float} 
\usepackage{subfigure} 

\title{GeoSegNet: Point Cloud Semantic Segmentation via Geometric Encoder-Decoder Modeling}

\author{Chen Chen\textsuperscript{1}, Yisen Wang\textsuperscript{1}, Honghua Chen\textsuperscript{1}, Xuefeng Yan\textsuperscript{1}, \\\textbf{Dayong Ren\textsuperscript{2}, Yanwen Guo\textsuperscript{2}, Haoran Xie\textsuperscript{3}, Fu Lee Wang\textsuperscript{4}, Mingqiang Wei\textsuperscript{1}}\\
\textsuperscript{1}Nanjing University of Aeronautics and Astronautics, Nanjing, China\\
\textsuperscript{2}Nanjing University, Nanjing, China\\
\textsuperscript{3}Lingnan University, HongKong, China\\
\textsuperscript{4}Hong Kong Metropolitan University, HongKong, China\\
\texttt{\{chenxx,eason,yxf,mqwei\}@nuaa.edu.cn}\\
\texttt{\{chenhonghuacn,rdyedu@\}@gmail.com} \\
\texttt{yw.guo@nju.edu.cn} \\
\texttt{hrxie@ln.edu.hk}\\
\texttt{pwang@hkmu.edu.hk}\\
}

\iclrfinalcopy
\begin{document}

\maketitle

\begin{abstract}
Semantic segmentation of point clouds, aiming to assign each point a semantic category, is critical to 3D scene understanding.
Despite of significant advances in recent years, most of existing methods still suffer from either the object-level misclassification or the boundary-level ambiguity. In this paper, we present a robust semantic segmentation network by deeply exploring the geometry of point clouds, dubbed GeoSegNet. Our GeoSegNet consists of a multi-geometry based encoder and a boundary-guided decoder. In the encoder, we develop a new residual geometry module from multi-geometry perspectives to extract object-level features. In the decoder, we introduce a contrastive boundary learning module to enhance the geometric representation of boundary points. Benefiting from the geometric encoder-decoder modeling, our GeoSegNet can infer the segmentation of objects effectively while making the intersections (boundaries) of two or more objects clear. Experiments show obvious improvements of our method over its competitors in terms of the overall segmentation accuracy and object boundary clearness. 
Code is available at \href{https://github.com/Chen-yuiyui/GeoSegNet} {https://github.com/Chen-yuiyui/GeoSegNet}.
\end{abstract}

\section{Introduction}
Semantic segmentation of point clouds is fundamental in parsing 3D scenes for, e.g., autonomous driving and augmented reality.
It seeks to assign an accurate category label to each point in the point cloud.
Currently, semantically segmenting point clouds of arbitrary complex scenes remains challenging yet not well solved, due to the nature of unbalanced point distribution and representation irregularity. 

Existing efforts of point cloud semantic segmentation are often projection-based \citep{boulch2017unstructured,lawin2017deep,milioto2019rangenet++,guo2020deep}, discretization-based \citep{maturana2015voxnet,graham20183d,riegler2017octnet,tchapmi2017segcloud} or point-based \citep{qi2017pointnet++,jiang2018pointsift,wang2019dynamic,thomas2019kpconv,hu2020randla}. Both projection- and discretization-based methods reduce the resolution of point clouds, inevitably leading to the loss of certain geometric information. 
Also, they possess a high computational complexity, due to the projection of point clouds to regular representations (e.g., 2D grid or 3D voxel) and the re-projection of results back to the point cloud domain. In contrast, point-based methods can deal with the irregularity of
point clouds without conversions, which leverage multi-layer perceptrons (MLP) to directly process each point and aggregate geometric features through a max-pooling operation \citep{qi2017pointnet++,jiang2018pointsift,wang2019dynamic,thomas2019kpconv,hu2020randla}.

\begin{figure}[ht]%
\begin{center}
    \includegraphics[width=1\textwidth]{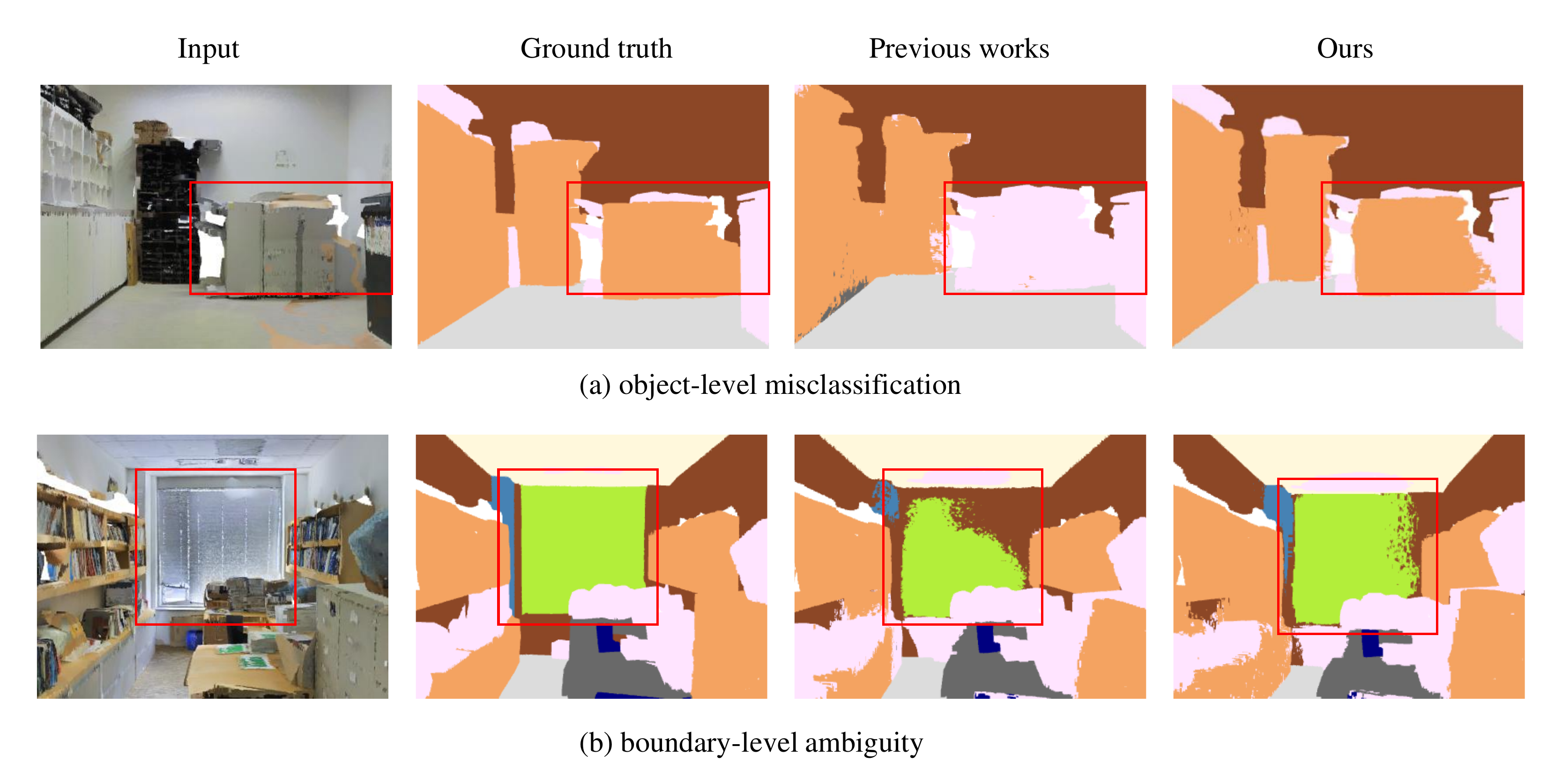}
\end{center}

\caption{Either the object-level misclassification or the
boundary-level ambiguity will appear when segmenting complicated 3D scenes. 
The cutting-edge point-based network \citep{xu2021investigate} is difficult to distinguish the two objects of different categories, thus merging them together. This leads to the object-level misclassification (upper). Meanwhile, it usually performs poorly on the object boundaries, leading to the ambiguous transition area between two objects (bottom). }\label{fig7}
\end{figure}

Although recent years have witnessed the success of point-based methods, they still suffer from the object-level misclassification or the boundary-level ambiguity. We show an example in Fig. \ref{fig7}. One can observe that, cutting-edge point-based networks are hard to distinguish the two objects of different categories, thus merging them together. This leads to the object-level misclassification. Meanwhile, these methods usually perform poorly on the object boundaries, leading to the ambiguous transition area between two objects.

There are mainly two reasons to degenerate the performance of cutting-edge segmentation models.
First, the point-based methods often extract point features based on PointNet/PointNet++.
However, 1) PointNet/PointNet++ abandons considerable local features during max-pooling. 2) The global context depicts the semantics of a whole scene and also the correlations between different objects in the scene; PointNet/PointNet++ ignores the global context and only extracts the high-level feature representation by continuously expanding the receptive field. 
Such two drawbacks of PointNet/PointNet++ are inherited by these point-based segmentation methods.
\textit{We attempt to preserve different-type yet beneficial geometric features, as well as exploit the complementary RGB color information, by designing a comprehensive residual geometry module.} 
Second, multiple objects exist in a single scene. The intersections between these objects form the boundaries \citep{gong2021boundary}.
when points are located at/around the boundary of two or more different categories of objects, the feature of such a point gathered by its local neighboring points will couple different categories of information. As the network goes deeper, these confusing features from the boundary area will propagate to more other points \citep{hu2020jsenet,gong2021boundary,tang2022contrastive}. Thus, these points on the boundary confuse the network, making it hard to clearly segment the boundary region. 
\textit{We attempt to explicitly enhance feature discrimination in boundary areas by contrasting their representations.}

Based on the above two observations, we propose a novel geometric encoder-decoder model, called GeoSegNet, which not only effectively exploits the geometric information of the whole object, but also explores the boundary area.
First, we propose a new residual geometry module, consisting of four blocks, including the eigenvalue block, the geometric context feature representation block, the color feature representation block, and the residual MLP block. Existing learning-based local geometric descriptors are sensitive to rigid transformations (i.e., translation, rotation), which are neither robust nor representative. 
Second, we utilize contrastive learning to focus on the regions of boundary points. 

We leverage the proposed residual geometry module to obtain geometric features that are robust to orientation and $z$-axis rotation, which are important to focus on the features of points in both the boundary and interior areas. Furthermore, we design a multi-stage loss function to make the contrastive boundary learning block learn more distinct features from different sub-sampled point clouds in each stage. Our GeoSegNet helps to learn the characteristics of both the boundaries and interiors more efficiently, as well as improving the overall performance.

The main contributions are summarized as follows:
\begin{itemize}
\item  We propose a novel point cloud segmentation network, dubbed GeoSegNet, which takes both the boundary contrastive learning and geometric representation into consideration, for more accurate segmentation.
\item  We devise a residual geometry module with $z$-axis rotation invariant, which better captures geometric context information, by encoding eigenvalues, relative angles, and distances, and color variances. 
\item We employ contrastive boundary learning (CBL) to improve the feature learning and representation of points in boundary areas.
\item We propose a multi-stage loss function to progressively enhance the feature descriptions of different sub-sampled point clouds, making the geometric features represent points better.
\end{itemize}

\section{Related Works}\label{sec2}
\subsection{Point Cloud Semantic Segmentation}
Point cloud semantic segmentation aims to assign semantic labels to each 3D point. Recent deep learning methods have replaced traditional methods that use manual features, which can be divided into three categories \citep{guo2020deep}: projection-based \citep{boulch2017unstructured,lawin2017deep,milioto2019rangenet++}, discretization-based \citep{maturana2015voxnet,riegler2017octnet,tchapmi2017segcloud} and point-based methods \citep{qi2017pointnet,qi2017pointnet++,zhao2019pointweb}. 

Projection-based methods project the 3D points to 2D images \citep{lawin2017deep,boulch2017unstructured}. For the 2D image plane, many previous works make use of the existing methods for 2D image semantic segmentation processing. However, the projection will cause the loss of information, which results in lower accuracy. Besides, these methods are computationally expensive due to the back projection. Discretization-based methods convert the point cloud into voxels, which are then fed into a 3D convolutional network to generate voxel-level labels. Many researchers \citep{maturana2015voxnet,tchapmi2017segcloud,graham20183d} achieved 3D point cloud segmentation by Discretization-based methods. However, these methods inevitably destroy geometric information due to voxelization and need intensive computation power.
Compared with the previous methods, point-based methods directly operate on 3D points, while the pioneering work is PointNet \citep{qi2017pointnet}. Then, PointNet++ \citep{qi2017pointnet++} introduced a hierarchical network to aggregate features from different scales. Following these great success, recent works adopt an standard encoder-decoder, and propose various local aggregation modules to better extract features,including attention-based aggregation \citep{guo2021pct,hu2020randla,zhao2021point}, graph-based aggregation \citep{wang2019dynamic,wang2019graph}, 3D-convolution \citep{boulch2020fkaconv,wu2019pointconv}, kernel-based aggregation \citep{thomas2019kpconv}. Despite these different modules all improving the performance of segmentation in varying degrees, the boundaries of point cloud segmentation are rarely explored. Compared with these methods, our method is boundary-inner aware, which could learn the characteristics of both the boundaries and the inner areas more eﬀiciently, as well as improve the overall performance.

\subsection{Boundary in Semantic Segmentation}
The boundary in point clouds usually refers to the area between two or more objects which belonging to different categories. JSENet \citep{hu2020jsenet} and Boundary-Aware GEM \citep{gong2021boundary} noticed the importance of boundary in 3D point cloud segmentation. These methods largely increase the complexity of the model, but have little improvement on the overall performance. Tang et al. \citep{tang2022contrastive} proposed a novel contrastive boundary learning (CBL) framework for point cloud segmentation, which achieves compelling performance on boundaries. However, CBL pays too much attention to the boundary area and does not extract the geometric features of the object more precisely. To this end, we propose an effective neural network, which not only extracts the spatial geometric context feature of the point cloud, but also focuses on the boundary points and enhances the feature discrimination across the boundaries. 

\section{Methods}\label{sec3}
In this section, we firstly present our GeoSegNet, which is built based on an encoder-decoder architecture incorporated with a residual geometry module and contrastive boundary learning (CBL) module. Then we introduce the details of our residual geometry module and contrastive boundary learning module. 
\subsection{Overview}\label{subsec4}
\begin{figure}[ht]%
\centering
\includegraphics[width=1.0\textwidth]{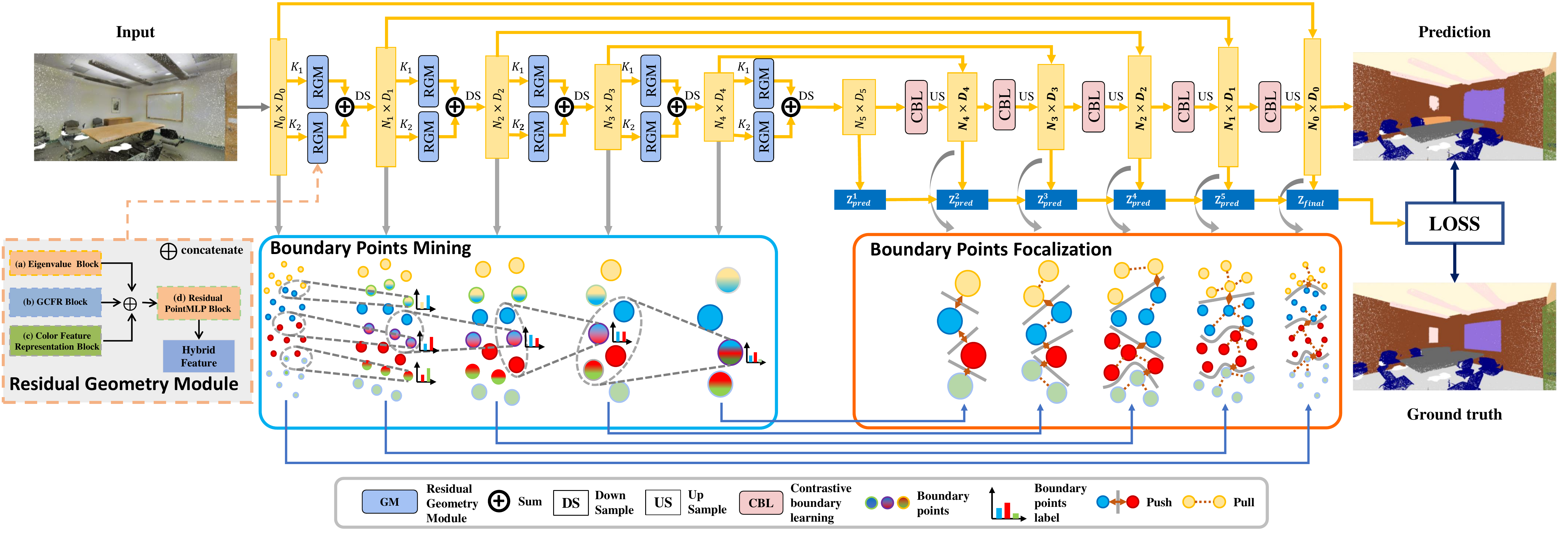}
\caption{The overall architecture of GeoSegNet. It mainly consists of a Residual Geometry Module (RGM) and a Contrastive Boundary Learning (CBL) Module for each layer. RGM can obtain features of each point from two different respective fields $K_{1}$ and $K_{2}$. At the same time, CBL is used to find the boundary points and enhance the feature discrimination of the boundaries. The blue and orange boxes below show the boundary points mining and boundary points focalization process in CBL. }\label{fig4}
\end{figure}

We embed the residual geometry module and CBL module into a widely adopted encoder-decoder architecture, and construct our segmentation network, GeoSegNet. At the top level, the architecture of GeoSegNet is depicted in Figure \ref{fig4}.

We denote the input point cloud as $P=\left \{ p_{i} \in R^{ N \times D_{0}},i=1,2,..., N  \right \}$, where $N$ is the number of points and $D_{0}$ denotes the dimension of $xyz$ coordinates and color information $(r, g, b)$. Five encoder stages are used to implement feature embedding step by step. Among them, the point cloud is sub-sampled from $N$ to $\frac{N}{256}$ by the farthest point sample method, and the embedded residual geometry module can obtain features of each point from two different respective fields $K_{1}$ and $K_{2}$. Then, we utilize a hierarchical propagation strategy with the nearest-neighbor interpolation in five decoder stages. It simply utilizes the distance of the nearest neighbors to obtain the up-sampled features and concatenates the up-sampled features with the encoder outputs through skip connections. We propagate points features to the next stage, and obtain predictions $Z_{pred}^{n}$ in each stage $n$. At the same time, the contrastive boundary learning module is used to find the boundary points and enhance the feature discrimination of the boundaries. The blue and orange boxes in Figure \ref{fig4} show the boundary points mining and boundary points focalization process in CBL. At last, two MLPs are used to get the final predictions $Z_{final}$. The size of final predictions is $N\times C$, where $C$ is the number of categories.

\subsection{Residual Geometry Module}\label{subsec1}
\begin{figure}[htbp]%
\centering
\includegraphics[width=1\textwidth]{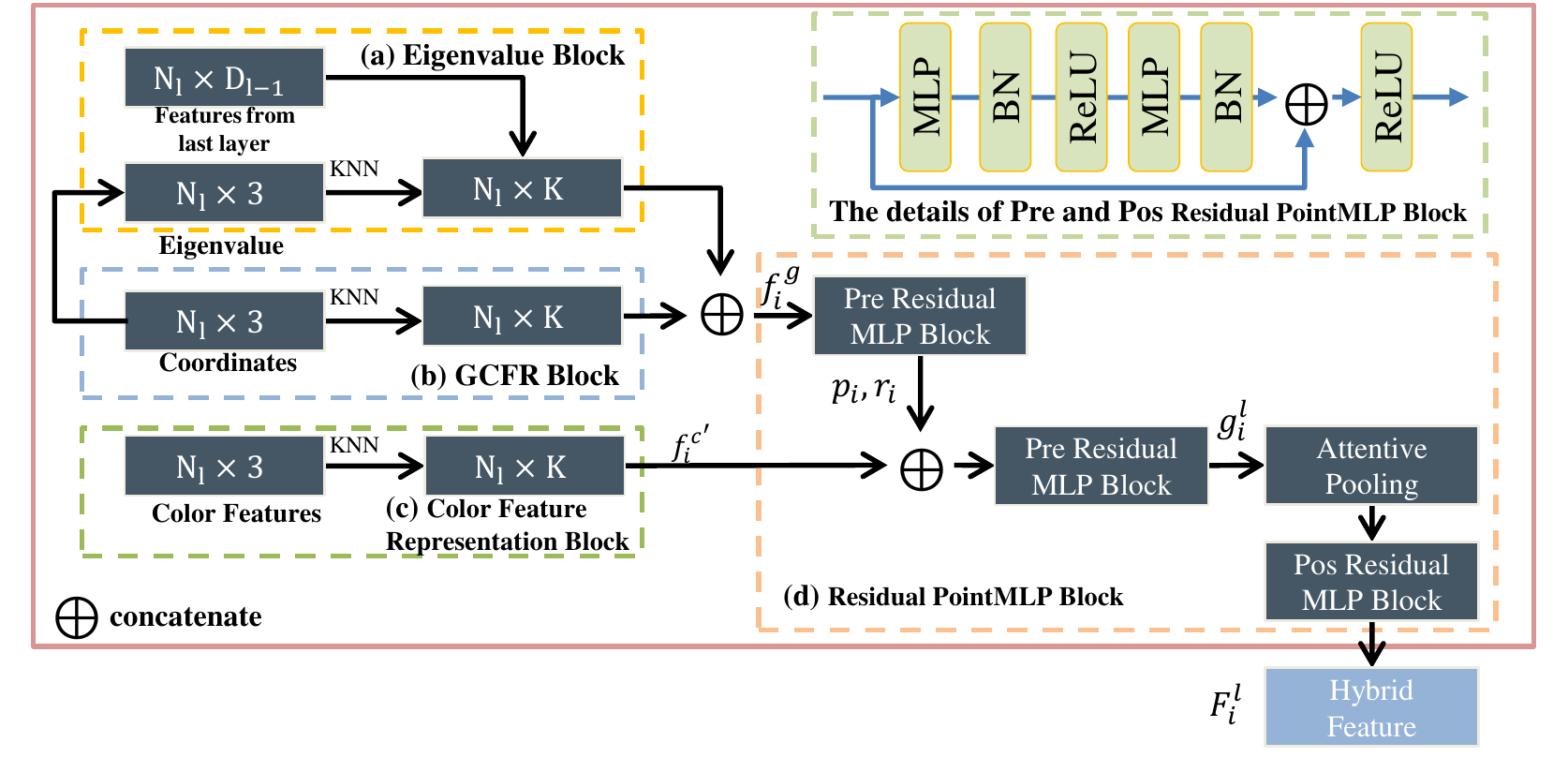}
\caption{The architecture of residual geometry module. The residual geometry module is constituted of four blocks: the
eigenvalue block, the geometric context feature representation block, the color
feature representation block, and the residual PointMLP block.}\label{fig1}
\end{figure}

The more effectively representing local features of the point cloud, the more accurately the network can distinguish the points in the boundary area. However, the features directly captured by the network have a limited capability to utilize the geometric information and are sensitive to rigid transformation. Therefore, we introduce the residual geometry module to obtain geometric features that are robust to rotation and orientation. As shown in Figure \ref{fig1}, the residual geometry module is constituted of four blocks: the eigenvalue block, the geometric context feature representation (GCFR) block, the color feature representation block, and the residual PointMLP block. We obtain the relative distance and angle of each point by the geometric context feature representation block, and we calculate the eigenvalues to describe each point better with the invariance of rotation and translation. Then, we use the color feature representation block to gather the color features and calculate the $K$-nearest neighboring variance of each point. After these blocks, we utilize the residual PointMLP block to mine deeper features and retain the original features.
\subsubsection{Eigenvalue Block}\label{subsubsec1}
In order to better describe each point with the invariance of rotation and translation, we calculate the eigenvalues from $K$-nearest neighbors of each point $p_{i}$ in the Euclidean space. Note that we only use the coordinates of each $p_{i}$ in this block. Let $\left\{x_{i_{1} },...,x_{i_{k}}\right\}$ be $K$-nearest neighbors of each point $p_{i}$. We denote $M = \left\{{x_{i_{1}}-p_{i}},...,x_{i_{k}}-p_{i}\right\}$, and compute a symmetric positive definite matrix $C=MM^\top$ \citep{xu2020geometry}. Then, by decomposing the matrix $C$, the eigenvalues $(\lambda _{i}^{1},\lambda _{i}^{2},\lambda _{i}^{3}) (1\leq i \leq N)$ and the corresponding eigenvectors $(v ^{1},v ^{2},v ^{3})$ are obtained respectively, which meet the following constraints:
\begin{equation}\label{eq1}
    Cv^{i} = \lambda^{i}v ^{i}, i\in \left\{1,2,3\right\}
\end{equation}

Let $R$ be an arbitrary rotation matrix in a 3D Euclidean space. By applying the rotation matrix to the point cloud, we obtain a new matrix $C ^{'} = RM(RM)^{\top}$. The inverse of a rotation matrix is its transpose matrix as:
\begin{equation}\label{RRTeq1}
    RR ^{T}=RR^{-1}=R^{\top}R =I
\end{equation}

Then, we can deduce that:
\begin{equation}\label{RRTeq2}
    C^{'}Rv^{i} = RMM ^{\top}R ^{\top }Rv ^{i} = RMM ^{\top }v ^{i}=RCv ^{i} = R \lambda^{i}v ^{i}
\end{equation}

It is easy to observe that the calculated eigenvalues are invariant to rotation and translation. Since the eigenvalue feature can represent the shape geometry to a certain extent, we then group the $K$-nearest neighbors for each point in the eigenvalue space. The feature of point $p_{i}$ in the eigenvalue space is defined as $\left \{ e_{i_{1} }^{l-1} ,...,e_{i_{k} }^{l-1}  \right \} $, where $l$ denotes the Eigenvalue Block in the $l$-th layer. As shown in Figure \ref{fig1}, the inputs of the first layer are the coordinates and the eigenvalues $(\lambda _{i}^{1},\lambda _{i}^{2},\lambda _{i}^{3})$. In the later layers, we use the output of the previous layer as the input features, and group features from both the Euclidean space and eigenvalue space.

%This is similar to DGCNN \citep{wang2019dynamic}'s idea, but we group the similar features both in Euclidean space and eigenvalue space. Unlike DGCNN, our method does not use dynamic strategy, we use eigenvalues decomposition to select the nearest neighbor and share local features with distant points with similar geometric information.
%In many real-world indoor scenes, the orientations of the same category object are  generally different, such as sofas and chairs. Since the boundaries of sofas and chairs have a certain curvature, the boundary is less helpful for segmentation. 
\subsubsection{Geometric Context Feature Representation Block}\label{subsubsec2}
In many real-world indoor scenes, the orientations and rotations of the same category object are very likely different, such as sofas and chairs. To enhance the the robustness to orientations and rotations of the learned features and capture long-distance context information, we design a geometric context feature representation block to employ different-type yet beneficial geometries to promote the feature representation.
Inspired by SCF-Net \citep{fan2021scf}, we use the polar representation to calculate the relative angle and distance between each point $p_{i}$ and the centroid of its $K$-nearest neighbors. Different from only using the Cartesian coordinate system, the relative angle and distance expressed by spherical coordinate system are invariant to z-axis rotation.

\begin{figure}[ht]%
\centering
\includegraphics[width=1\textwidth]{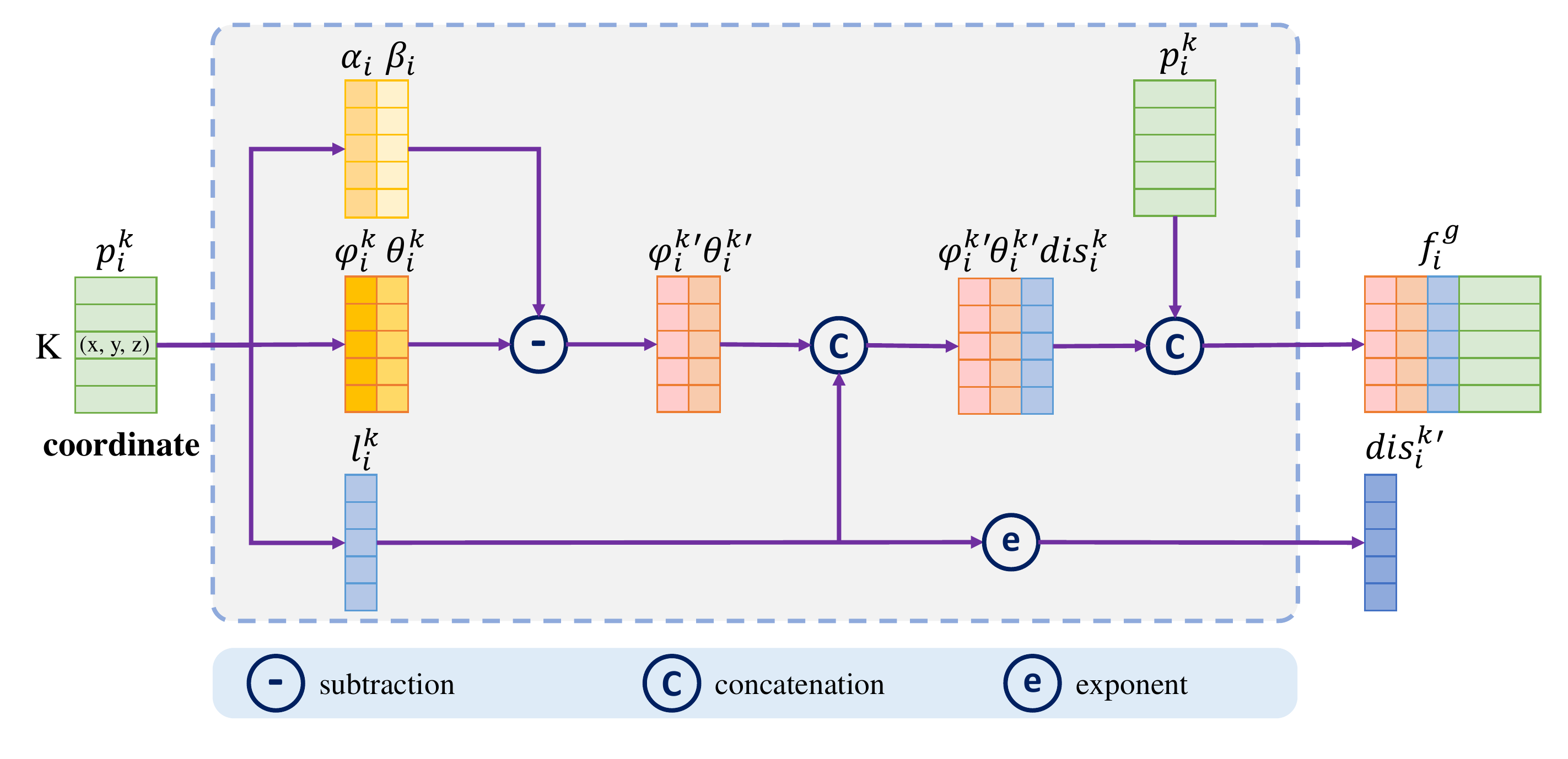}
\caption{ Geometric Context Feature Representation Block.}\label{fig2}
\end{figure}

As shown in Figure \ref{fig2}, the local spatial information is fed into the GCFR block, and the output is the angle and geometric distance. The detailed calculation steps of GCFR Block are as follows:

\textbf{Calculating distance and relative angle.}
Firstly, we calculate the geometric distance $dis_{i}^{k}$ and the original relative angles $\theta_{i}^{k}$, $\phi_{i}^{k} $. Let $\left \{  p_{i_{1} },...p_{i_{k}},...,p_{i_{K} } \right \} $ be $K$-nearest neighbors of each point $p_{i}$. 
\begin{equation}\label{eq3dis}
    dis_{i}^{k} = \sqrt{x_{i}^{k^{2} }+ y_{i}^{k^{2} }+z_{i}^{k^{2} }}  
\end{equation}

\begin{equation}\label{eq4angle1}
    \phi_{i}^{k} = \arctan \frac{y_{i}^{k}}{x_{i}^{k} }
\end{equation}

\begin{equation}\label{eq4angle2}
    \theta_{i}^{k} = \arctan \frac{z_{i}^{k}}{\sqrt{x_{i}^{k^{2}}+ y_{i}^{k^{2}}}}
\end{equation}
where $(x_{i}^{k}, y_{i}^{k}, z_{i}^{k})$ is the coordinate in the Cartesian coordinate system.

\textbf{Calculating the angle of centroid.}
Then, we calculate the centroid $p_{i}^{c}$ of the $K$-nearest neighbors. We define the direction from $p_{i}$ to $p_{i}^{c}$ as a local direction with two benefits. Firstly, the centroid can reflect the overall characteristics of $K$-nearest points. Secondly, by using the centroid of each $p_{i}$, we can effectively reduce the uncertainty of down sampling. We follow Equations (\ref{eq4angle1}) and (\ref{eq4angle2}) to calculate the relative angles of $p_{i}^{c}$, which are defined as $\alpha_{i}^{c}$ and $\beta_{i}^{c}$.

\textbf{Calculating the relative angle.}
After calculating the angle of centroid, we can get the final relative angle $\phi_{i}^{k^{'}}$ and $\theta_{i}^{k^{'}}$, respectively as:
\begin{equation}\label{eq7angle1}
    \phi_{i}^{k^{'}} = \phi_{i}^{k} - \alpha_{i}^{c}
\end{equation}
\begin{equation}\label{eq8angle2}
    \theta_{i}^{k^{'}} = \theta_{i}^{k} - \beta_{i}^{c}
\end{equation}

\textbf{Calculating the local density.}
The local point cloud density, as an internal attribute of point cloud, is related to the point cloud interval. A sharp change in density usually indicates a change in the object to which the point belongs. In order to obtain more geometry features as a supplement, we propose the \textit{Local Density} to learn some features of the global context. We define $r_{i} =\frac{v_{k}}{v_{g}}$  that is the the volume ratio of local and global bounding spheres, where $v_{k}$ is volume of bounding sphere of K-nearest neighbors of $p_{i}$, and $v_{g}$ is the volume of the bounding sphere of the input point cloud. We calculate the radius of the local sphere by calculating the farthest $dis_{i}^{k}$ from of each $p_{i}$ in its $K$-nearest neighboring points. It represents the density of a local region.

\begin{figure}[ht]%
\centering
\includegraphics[width=1\textwidth]{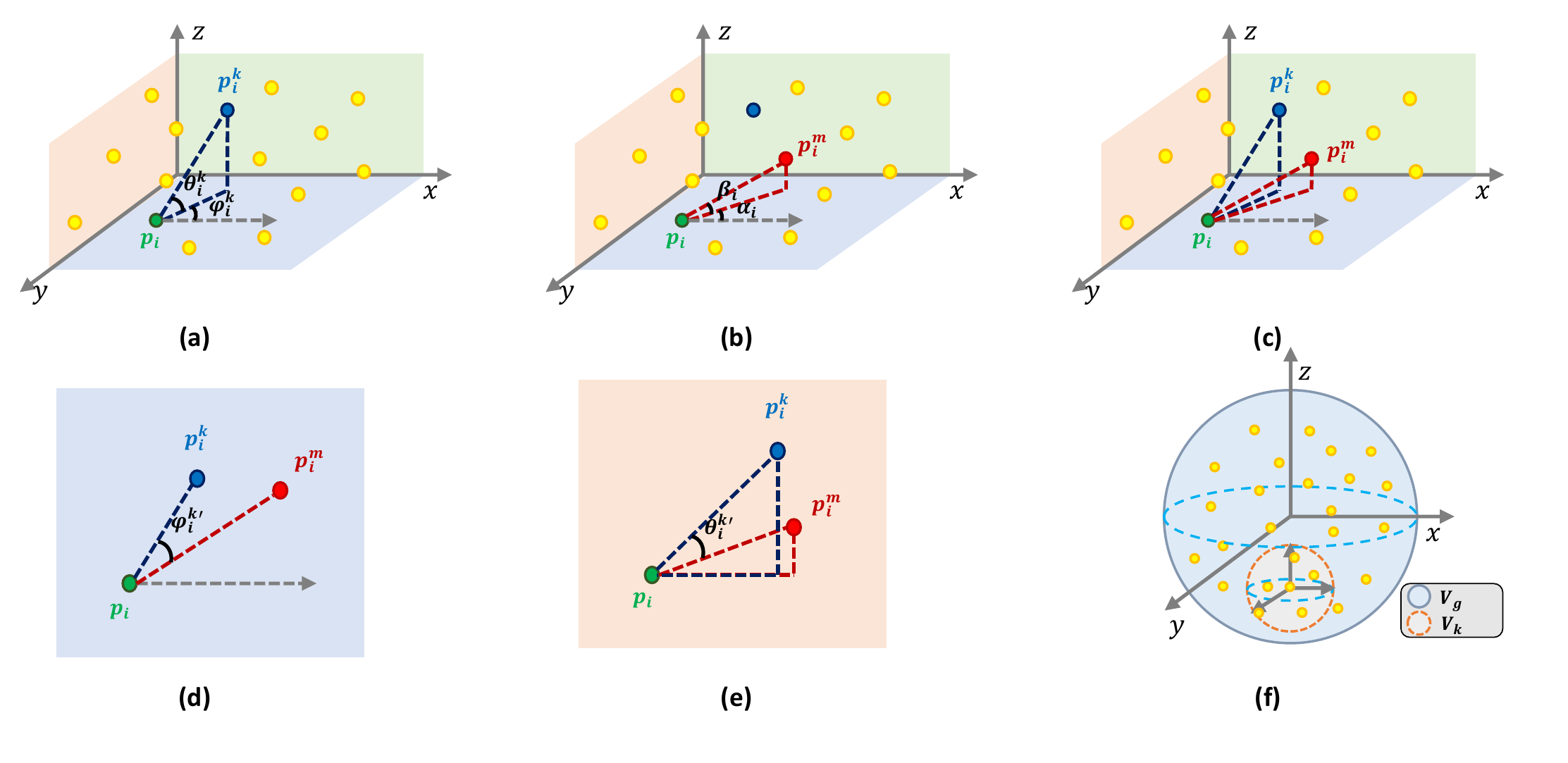}
\caption{The steps of Geometric Context Feature Representation Block. (a) Calculating distance and relative angle $\theta_{i}^{k}, \phi_{i}^{k} $; (b) Calculating the angle of centroid $\alpha_{i}^{c}$ and $\beta_{i}^{c}$;
(c,d,e) Calculating the relative angle $\phi_{i}^{k^{'}}$ and $\theta_{i}^{k^{'}}$. (f) Calculating the local density. }\label{fig3}
\end{figure}

The above computed geometric representations are illustrated in Figure \ref{fig3}. We concatenate the output of GCFR block and eigenvalue block so that the local representation is invariant to the $z$-axis rotation, defined as $f_{i}^{g}$.

\subsubsection{Color Feature Representation Block}\label{subsubsec3}
Color information and geometry information are complementary. If points have similar color features and geometric features, there is a high probability that they belong to the same category. To capture typical color features, we propose the color feature representation block. It takes the color feature variance of the $K$-nearest neighborhood as the aggregation feature of each point $p_{i}$. Let $\left\{f_{i_{1}},...,f_{i_{k}}\right\}$be $K$-nearest neighbors’ RGB-features of point $p_{i}$, and $f_{i}$ is the RGB-feature of $p_{i}$. We group the neighboring features and calculate their variance as follows:
\begin{equation}\label{eq8}
    f_{i}^{k} =   ((f_{j} - f_{i})\oplus f_{i}), j \in \left\{ i_{1},...,i_{k}\right \}
\end{equation}
\begin{equation}\label{eq9}
    f_{i}^{\sigma} =\frac{\left(f_{i}-f_{i_{1}}\right)^{2}+\cdots+\left(f_{i}-f_{i_{k}}\right)^{2}}{K}
\end{equation}
where $\oplus$ means the concatenation operation. We concatenate $f_{i}^{k}$ with $f_{i}^{\sigma}$ as the color feature representation $f_{i}^{c'}$ for each point $p_{i}$:
\begin{equation}\label{eq10}
    f_{i}^{c'} = (f_{i}^{k}\oplus f_{i}^{\sigma})
\end{equation}

By using the variance of color feature, we can effectively enhance the representation of each point, which is helpful to distinguish the boundary points.
\subsubsection{Residual PointMLP Block}\label{subsubsec4}
Ma et al. \citep{ma2022rethinking} propose a simple and effective pure residual MLP network for point cloud analysis, called PointMLP, which does not integrate complex local geometry extractors, but still performs competitively. Inspired by PointMLP \citep{ma2022rethinking}, we simplify their residual modules .

The key operations can be formulated as:
\begin{equation}\label{resmlp}
        g_{i}^{l} =R_{pre}(R_{pre} (f_{i}^{g})\oplus p_{i}\oplus r_{i}\oplus f_{i}^{c'})
\end{equation}
\begin{equation}\label{resmlp2}
          F_{i}^{l} = R_{pos}(Attention(g_{i}^{l}))
\end{equation}
where $R_{pre}(\cdot)$ and $R_{pos}(\cdot)$ are residual point MLP blocks. The residual MLP block is combined by fully connected layers, normalization layers and activation layers. Each layer repeats twice as shown in Figure \ref{fig4}. $f_{i}^{c'}$ is the color feature from the Color Feature Representation Block, $f_{i}^{g}$ is the geometric context feature, and $r_{i}$ is the local density. 

$Attention(\cdot)$ means that we use the attentive pooling to aggregate the set of neighboring point features. Previous works \citep{qi2017pointnet++,li2018pointcnn} use max or average pooling to integrate the neighboring features, which results in the loss of information. To this end, we take an attentive pooling \citep{hu2020randla} :
\begin{equation}\label{attentive}
         \tilde{g_{i}^{l}} =  {\textstyle \sum_{k=1}^{K}mlp(g_{i}^{l}[k])\cdot{g_{i}}^{l}[k] } 
\end{equation}
Moreover, as Figure \ref{fig4} shows, we utilize the residual geometry module to obtain features of each point $p_{i}$ from two different respective fields, which can further enhance the representation of each point $p_{i}$. We choose $K_{1}$ and $K_{2}$ nearest neighbors for the residual geometry module, and get the final output of the $l$-th encoder layer which denoted as $F^{l} = F_{K_{1}}^{l} + F_{K_{2}}^{l}$.

\subsection{Contrastive Boundary Learning Module}\label{subsec2}
\subsubsection{Boundary Points Mining}\label{subsubsec6}
To better obtain the boundaries of point clouds, we examine the boundaries in each sub-sampled point cloud at multi-scales, which makes it possible to use the contrastive boundary learning on each sub-sampled stage. We can easily collect the boundary points from the input point cloud and define the boundary points $B_{i}$:
\begin{equation}\label{boundary}
    B_{i} = \left\{ p_{i} \in R \mid \exists \ p_{j}\in N_{i},  l_{j}\neq l_{i}\right\}
\end{equation}
where $N_{i}$ is the local neighborhood of $p_{i}$ with a radius of 0.1, following the common practice \citep{thomas2019kpconv,liu2020closer}. $l_{i}$ is the ground-truth label of $p_{i}$.

However, the label of the sub-sampled point cloud is undefined due to sub-sampling. It is difficult to obtain a correct definition of boundary points following Eq.(\ref{boundary}). To this end, we propose the boundary points mining which could determine the ground-truth boundary points in each sub-sampled stage. We define the sub-sampled point cloud as $R^{n}$, where $n$ is the stage. For the original input point cloud, we have $R^{0} = R$. When collecting the boundary points in stage $n$, it is necessary to determine the label $l_{i}^{n}$ of each sub-sampled point $p_{i}^{n}\in R^{n}$. We utilize the sub-sampling procedure to determine the label, and we have the following:
\begin{equation}\label{boudarymining}
    l_{i}^{n} = \mathrm{MEAN} (\left\{ l_{j}^{n-1} \mid \ p_{i}^{n-1} \in N^{n-1}(p_{i}^{n-1})\right\})
\end{equation}
where $N^{n-1}(p_{i}^{n})$ denotes the neighbors of $p_{i}^{n}$ in previous stage, and MEAN$(\cdot)$ is the mean-pooling.
With the ground-truth labels and Eq.(\ref{boudarymining}), we can obtain the sub-sampled point  annotation $l_{i}^{n}$ as a distribution \citep{tang2022contrastive}. The $k$-th location describes the proportion of $k$-th category in the original input point cloud. Then we utilize $max(l_{i}^{n})$ to obtain the boundary points in Eq.(\ref{boundary}), and use the feature of each sub-sampled point for the boundary focalization. 
\subsubsection{Boundary Points Focalization}\label{subsubsec7}
As Figure \ref{fig4} shows, we use the contrastive boundary learning to enhance the feature discrimination across boundaries. After the boundary points mining, we can annotate the sub-sampled point cloud to discover the boundary points in each sub-sampling point cloud for contrastive boundary learning. 

We follow the InfoNCE loss and CBL \citep{tang2022contrastive} to define the contrastive optimization goal on boundaries. It could encourage the network to learn the representation more similarly to the same category from its neighbor points, and more differently from other category points. It is noted that we only use contrastive boundary learning on the boundary points:
\begin{equation}\label{CBL}
    L_{CBL}= \frac{-1}{\mid B_{i}\mid} \sum_{p_{i} \in B_{i}} \log \frac{\sum_{p_{j} \in {N}_{i} \wedge l_{j}=l_{i}}^{} \exp \left(-d\left(F_{i}, F_{j}\right) / \tau\right)}{\sum_{p_{k} \in {N}_{i}}^{} \exp \left(-d\left(F_{i}, F_{k}\right) / \tau\right)}
\end{equation}
where $F_{i}$ is the feature of $p_{i}$, $d(\cdot,\cdot)$ is L2-distance, and $\tau$ is the temperature in contrastive learning. First, we consider all the boundary points $B_{i}$ defined in Eq.(\ref{boundary}) from the ground truth of original input point clouds. Then, for each boundary point $p_{i}\in B_{i}$, we restrict the sampling of its positive and negative  samples to be within its $K$-nearest neighbors $N_{i}$. In this way, we obtain the positive samples of $p_{i}$ as $\left\{ p_{j} \in {N}_{i} \wedge l_{j}=l_{i}\right\}$, and negative samples as $\left\{ p_{j} \in {N}_{i} \wedge l_{j}\neq l_{i}\right\}$. Consequently, we can focus on the boundary points and enhance the feature discrimination across the boundaries, which is effective in improving the performance of segmentation on the boundary areas.

\subsection{Multi-stage Loss Function}\label{subsec3}
In order to refine the features of the boundary areas progressively, and make more distinguishing characteristics on different sub-sampled point clouds learned by the network, we apply the multi-stage loss as follows:
\begin{equation}
     L_{pred}^{n} = L_{CE}(Z_{GT}^{n}, Z_{pred}^{n})
\end{equation}
where $Z_{GT}$ is the ground-truth labels in the $n$ stage, and\textit{ CE} is the Cross Entropy loss. We use $\mathit{Softmax}$ to obtain the label prediction $z_{pred}^{n}$ probability of point $p_{i}$ in the $n$-th stage.
The final loss for training is as follows:
\begin{equation}
    Loss = L_{final} + \lambda_{1}(\sum_{n=1}^{5}L_{pred}^{n}) + \lambda_{2} L_{CBL}
\end{equation}
where $L_{final}$ is the Cross Entropy loss between final output predictions $Z$ and ground truth labels $Z_{GT}$. $\lambda_{1}$ and $\lambda_{2}$ are the loss weights.

\section{Experiments}\label{sec4}
The experiments can be divided into two parts. We demonstrate the performance of our method and compare it with the state-of-the-art methods on S3DIS Area-5 and 6-fold cross validation for the scene semantic segmentation task. Then, intensive ablation studies are conducted. 
\subsection{S3DIS Indoor Scene Segmentation}
\subsubsection{Dataset}
To prove the effectiveness of our proposed method, we now introduce various experiments on the S3DIS \citep{armeni20163d} dataset. S3DIS is a high-quality indoor scene dataset that we can better define the boundary points. It contains 3D RGB point clouds in 6 indoor areas covering 272 rooms. Each point is annotated with a corresponding label from 13 categories (wall, floor, chair, etc.). Following Boulch et al. \citep{boulch2020convpoint}, we report the results under two settings: testing on Area 5 and 6-fold cross validation. At the training time, we randomly select points in the considered point cloud, and extract all the points in an infinite vertical column (section is 2 meters) centered on this point. For each column, we randomly select 4096 points as the input to the network.  
\subsubsection{Implementation Details}
The proposed GeoSegNet is implemented by Pytorch. The main architecture is shown in Figure \ref{fig4}. Both the  $xyz$ coordinates and $rgb$ colors are used as inputs. In our experiments, we take IAF-Net \citep{xu2021investigate} as our baseline. IAF-Net selects indistinguishable points adaptively and enhances fine-grained features for points, especially those indistinguishable points. For training, we follow the setup of baseline: train by 100 epochs and set a batch size of 8 with 1 RTX 2080Ti. We set the loss weight $\lambda_{1} = 0.1$ and $\lambda_{2} = 0.2$, and the temperature $\tau$ in contrastive learning is 1. In addition, we use z-axis rotation, random scale, random jitter, and random drop color as data augmentation.
\subsubsection{Results}
We make comparisons with recent state-of-the-art semantic segmentation methods. Tables \ref{tab1} and \ref{tab2} show the results on S3DIS Area 5 and 6-fold cross-validation. We adopt mIoU, OA and mACC as the evaluation metric. 
\begin{figure}[htbp]%
\centering
\includegraphics[width=1\textwidth]{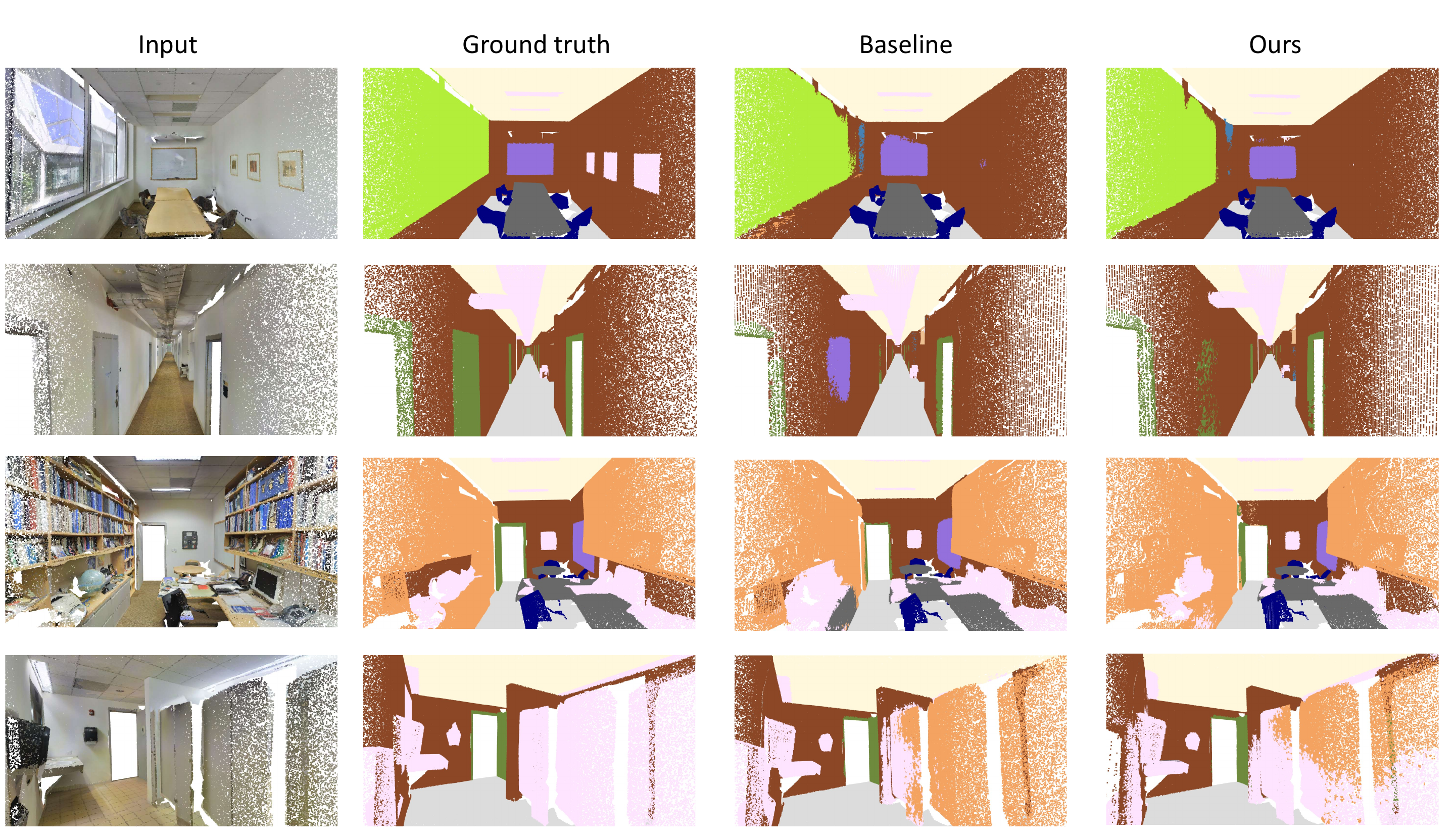}
\caption{Semantic segmentation results on S3DIS Area 5. From the left to the right is: input, ground truth, baseline (IAF-Net), ours (GeoSegNet).}\label{fig5}
\end{figure}
\begin{figure}[htbp]%
\centering
\includegraphics[width=1\textwidth]{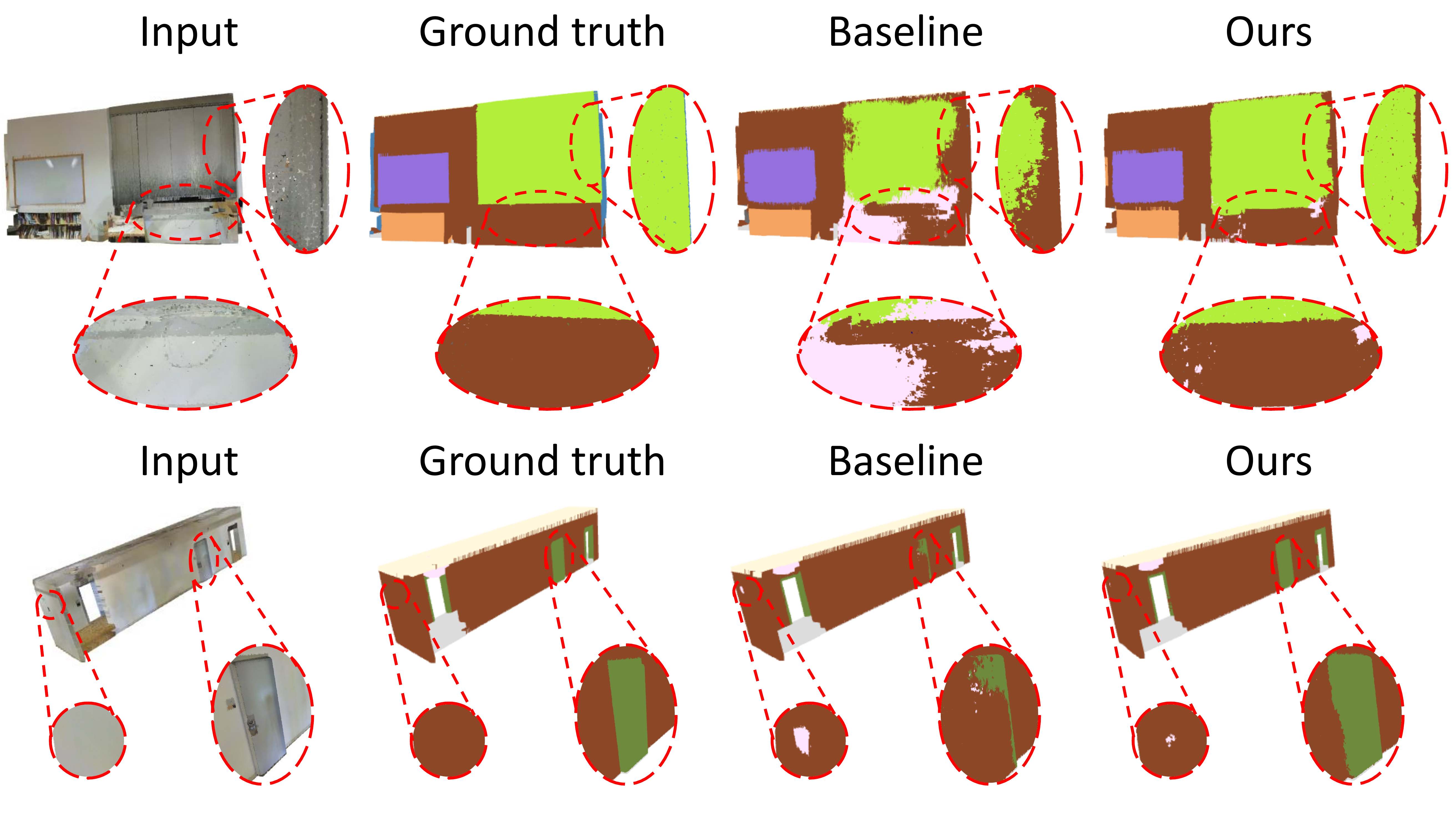}
\caption{Semantic segmentation results on boundary areas of S3DIS Area 5. From the left to the right is: input, ground truth, baseline (IAF-Net), ours (GeoSegNet). }\label{fig6}
\end{figure}

\begin{table}[ht]
\begin{center}
\caption{Semantic segmentation results on S3DIS evaluated on Area 5, shows the mean IoU (mIoU), overall accuracy (OA) and the mean accuracy (mACC). The classwise metric is IoU(\%). The best performance is highlighted in \emph{\textbf{bold}}, and the sub-optimal performance is marked by \emph{\underline{underline}}. }\label{tab1}%
\resizebox{\linewidth}{!}{
\begin{tabular}{l|ccc|ccccccccccccc}
\toprule
methods & mIoU  & OA & mACC & ceiling & floor & wall& beam &column &window &door &table &chair & sofa &bookcase &board &clutter\\
\midrule
PointNet \citep{qi2017pointnet}   & 41.1   & -   & 49.0 & 88.8 & 97.3 & 69.8 &\underline{0.1} & 3.9 & 46.3 & 10.8 & 59.0 & 52.6 & 5.9 & 40.3 & 26.4 & 33.2 \\
SegCloud \citep{tchapmi2017segcloud}   & 48.9  & –  & 57.4 &90.1 &96.1 &69.9 &0.0 &18.4& 38.4 &23.1& 70.4 &75.9 &40.9& 58.4& 13.0& 41.6  \\
PointCNN \citep{li2018pointcnn}    & 57.3 & 85.9& 63.9& 92.3& 98.2& 79.4& 0.0& 17.6& 22.8& 62.1& 74.4& 80.6& 31.7& 66.7 &62.1& \textbf{56.7}  \\
SPGraph \citep{landrieu2018large}  &58.0 &86.4 &66.5 &89.4 &96.9 &78.1 &0.0 &42.8 &48.9 &61.6 &84.7 &75.4 &69.8 &52.6 &2.1 &52.2\\
PCCN \citep{wang2018deep} &58.3 &- &67.0 &92.3 &96.2 &75.9 &\textbf{0.3} &6.0 &\textbf{69.5} &\underline{63.5} &66.9 &65.6 &47.3 &68.9 &59.1 &46.2\\
PointWeb \citep{zhao2019pointweb} &60.3 &87.0 &66.6 &92.0 &\textbf{98.5} &79.4 &0.0 &21.1 &59.7 &34.8 &76.3 &\textbf{88.3} &46.9 &69.3 &64.9 &52.5\\
PCT \citep{guo2021pct} &61.3 &- &67.7 &92.5 &\underline{98.4} &80.6 &0.0 &19.4 &61.6 &48.0 &76.6 &85.2 &46.2 &67.7 &67.9 &52.3\\
HPEIN \citep{jiang2019hierarchical} &61.9 &87.2 &68.3 &91.5 &98.2 &81.4 &0.0 &23.3 &65.3 &40.0 &75.5 &\underline{87.7} &58.5 &67.8 &65.6 &49.4\\
PointASNL \citep{yan2020pointasnl}&62.3 &87.7 & 68.5 &\textbf{94.3} &\underline{98.4} &79.1 &0.0 &26.7 &55.2 &\textbf{66.2} &\underline{83.3} &86.8 &47.6 &68.3 &56.4 &52.1\\
RandLA-Net \citep{hu2020randla} &62.4 &87.2 &\textbf{71.4} &91.1 &95.6 &80.2 &0.0 &24.7 &62.3 &47.7 &76.2 &83.7 &60.2 &71.1 &65.7 &53.8\\

GACNet \citep{wang2019graph} &62.9 &87.8 &- &92.3 &98.3 &81.9 &0.0 &20.4 &59.1 &40.9 &\textbf{85.8} &78.5 &\underline{70.8} &61.7 &\textbf{74.7} &52.8\\
Point2Node \citep{han2020point2node}&63.0 &\textbf{88.8} &70.0 &\underline{93.9} &98.3 &\textbf{83.3} &0.0 &\textbf{35.7} &55.3 &58.8 &79.5 &84.7 &44.1 &71.1 &58.7 &\underline{55.2}\\
SCF-Net \citep{fan2021scf}&\underline{63.4} &86.6 &71.2 &91.5 &95.0 &80.6 &0.0 &\underline{32.8} &60.0 &36.9 &71.9 &87.6 &\textbf{76.0} &\textbf{71.5} &\underline{69.9} &50.4\\
\bottomrule
IAF-Net \citep{xu2021investigate} &62.7 &87.9 &69.2 &93.7 &\textbf{98.5} &80.7 &0.0 &19.8 &49.6 & 53.8 &77.3 &86.8 &54.2 &69.9 &66.1 &54.5\\
Ours (GeoSegNet) &\textbf{64.9} &\underline{88.5} &\underline{71.3} &92.7 &98.1 &\underline{82.4} & 0.0 &31.2 &\underline{62.4} &62.6 &77.4 &87.1 &57.7 &\underline{71.2} & 67.3 &54.4\\
\bottomrule
\end{tabular}
}
\end{center}

\end{table}

\begin{table}[ht]
\begin{center}
\caption{Semantic segmentation results on S3DIS evaluated
on 6-fold cross-validation. The best performance is highlighted in \emph{\textbf{bold}}, and the sub-optimal performance is marked by \emph{\underline{underline}}.}\label{tab2}%
\begin{tabular}{l|ccc}
\toprule
methods & mIoU  & OA & mACC \\
\midrule
PointNet \citep{qi2017pointnet}   & 47.8   & 78.5 & 66.2 \\
SPGraph \citep{landrieu2018large}  &62.1 &86.4 &73.0 \\
DGCNN \citep{wang2019dynamic} &56.1 &84.1 &-\\
PointASNL \citep{yan2020pointasnl}&62.3 &87.7 & 68.5\\
PointCNN \citep{li2018pointcnn}    & 65.4 &\underline{88.1}& 75.6 \\
PointWeb \citep{zhao2019pointweb} &66.7 &87.3 &76.2\\
%KPConv \citep{thomas2019kpconv} &\textbf{70.6} &- &\underline{79.1}\\
%SCF-Net \citep{fan2021scf} &\textbf{71.6} &\textbf{88.4} %&\textbf{82.7}\\
RandLA-Net \citep{hu2020randla} &70.0 &88.0 &\textbf{82.0} \\	
\bottomrule
Ours(GeoSegNet) &\textbf{70.1} &\textbf{88.4} &\underline{76.9}\\
\bottomrule
\end{tabular}
\end{center}
\end{table}

\begin{figure}[htbp]%
\centering
\includegraphics[width=1\textwidth]{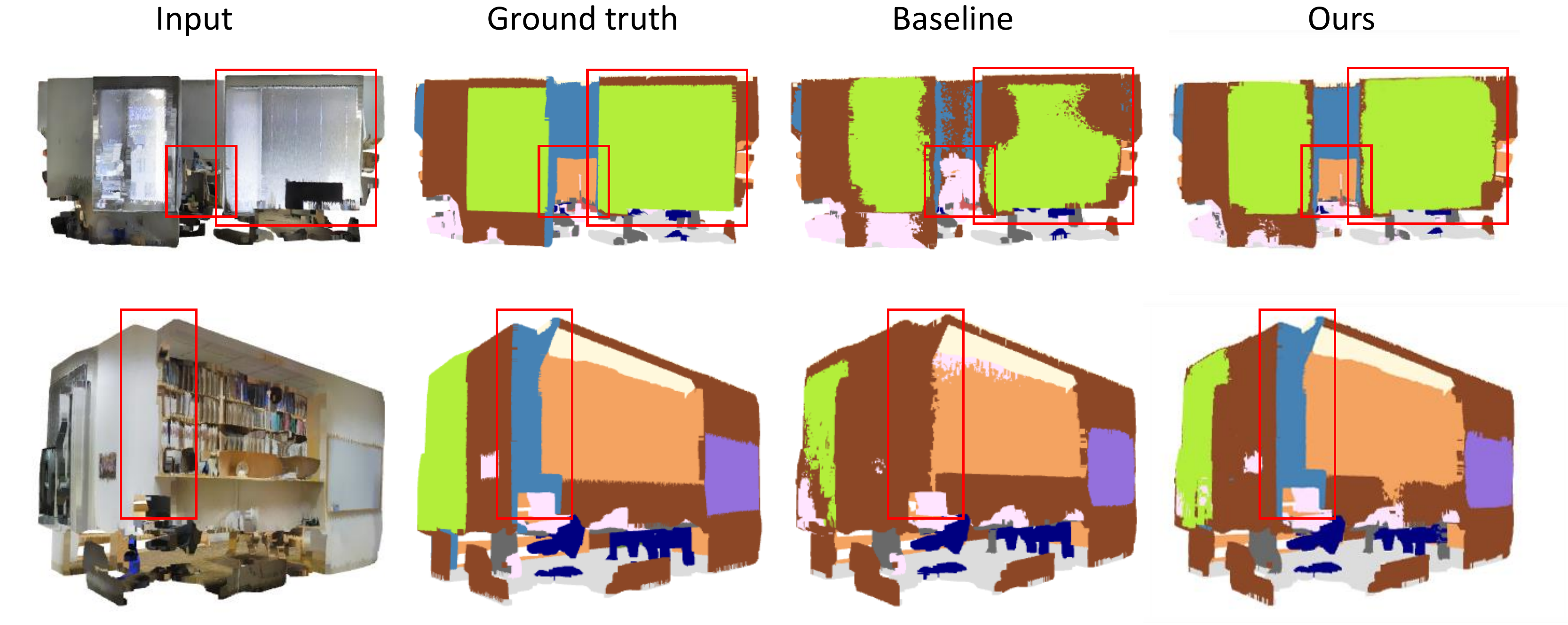}
\caption{Semantic segmentation results on boundary areas of S3DIS Area 5. From the left to the right is: input, ground truth, baseline (IAF-Net), ours (GeoSegNet). }\label{fig8}
\end{figure}

Table \ref{tab1} shows that our method performs better than the others on mIoU, and achieves comparable performance on OA and mACC. We achieve 64.9\% in mIoU on this benchmark which has
better performance than many state-of-the-art competitors. Compared with the baseline (IAF-Net), our method achieves better results in most categories, especially in those categories with distinct boundaries such as column, window, door. Due to the large curvature of the boundaries of the chair and sofa, it is difficult to learn boundary features from those categories. As Table \ref{tab1} shows, those categories are better segmented with the help of our residual geometry module, which makes those local representations invariant to the z-axis rotation. The results of IAF-Net are obtained by the official open-source code. Also, we visualize our segmentation results in Figures \ref{fig7}, \ref{fig5}, \ref{fig6} and \ref{fig8}. Misclassification easily appears in the transition region between two adjacent objects. For example, in the first row, points in ``wall'' category are predicted as ``board'' which is neighboring with wall. By contrast, the board's boundary of our method is significantly better than the baseline. In the third row, the segmentation results of chairs obtained by our method have better boundary contour thanks to our geometric-aware boundary contrastive learning network. We show more improvements in the details of the boundary area in Figure \ref{fig6}, which shows the effectiveness of our method for boundary areas.

To avoid overfitting on S3DIS Area 5, we further evaluate the S3DIS dataset on the 6-fold
cross-validation, with the result reported in Table \ref{tab2}. As can be seen in Table \ref{tab2}, our method achieves comparable performance (70.1\%) with the state-of-the-art method. 
\subsection{Ablation Study}
In this section, we conduct ablation studies to evaluate the effectiveness of each block in our method. We use the A3DIS Area 5 to evaluate the ablated networks, and show the result in Table \ref{tab3}. We conduct these studies on Area 5 of S3DIS with the same network parameters, using the proposed full GeoSegNet as best performance reference. 

We can see from Table \ref{tab3}, removing eigenvalue block and GCFR block decreases the performance by nearly 6.2\% mIoU. The decrease from the first row to the second row demonstrates that the introduction of local spatial geometric context feature can effectively improve performance. The decrease confirms the importance of the local spatial geometric context feature. Secondly, we only remove the GCFR block, the result shows that the effectiveness of the GCFR block. The result of removing CBL shows that boundary areas are worth attention, focusing on boundary points can gain a significant improvement. When we remove the multi-stage loss function, the performance is reduced as expected because of the multi scale features can enhance the encoding points representation.

\begin{table}[ht]
\begin{center}
\caption{The contribution of each block. Ablation studies on S3DIS Area 5 validation based on our full network. The best performance is highlighted in \emph{\textbf{bold}}, and the sub-optimal performance is marked by \emph{\underline{underline}}.}\label{tab3}%
\begin{tabular}{l|c}
\toprule
\ &mIoU(\%) \\
\midrule
The Full framework &\textbf{64.9}\\
removing Eigenvalue Block \& GCFR Block &58.7\\
removing GCFR Block &59.9\\
removing Residual MLP Block &64.4\\
removing Color Feature Representation Block &\underline{64.7}\\
removing Contrastive Boundary learning &63.6\\
removing Multi-stage Loss Function &64.1\\
\bottomrule
\end{tabular}
\end{center}
\end{table}

\begin{table}[ht]
\begin{center}
\caption{The effect of loss weight. The best performance is highlighted in \emph{\textbf{bold}}.}\label{tab4}%
\begin{tabular}{c|ccc}
\toprule
loss weight &mIoU &OA &mACC \\
\midrule
$\lambda_{1}$=0.1, $\lambda_{2}$=0.1 &64.6 &88.2 &71.1\\
$\lambda_{1}$=0.1, $\lambda_{2}$=0.2 &\textbf{64.9} &\textbf{88.5} &\textbf{71.3}\\
$\lambda_{1}$=0.1, $\lambda_{2}$=0.3 &64.0 &88.0 &\textbf{71.3}\\
\bottomrule
\end{tabular}
\end{center}
\end{table}

We conduct empirical study on S3DIS Area 5 to analyze the effect of loss weight $\lambda_{1}$ and $\lambda_{2}$ in the GeoSegNet. As shown in Table \ref{tab4}, we find that the
proper loss weight is $\lambda_{1}$=0.1, $\lambda_{2}$=0.2. 

\section{Conclusion}\label{sec5}
We have introduced, GeoSegNet, a new geometric encoder-decoder  designed for point cloud segmentation network,  which not only effectively exploits the geometric information of the whole object, but also explores the boundary area. GeoSegNet tries to preserve different-type yet beneficial geometric features, as well as exploit the complementary RGB color information, by a comprehensive residual geometry module. Besides, it explicitly enhances feature discrimination in boundary areas by contrasting their representations. Extensive experiments demonstrated the effectiveness of our method, showing that it outperforms the state-of-the-arts in various configurations.

\section{Declarations}

\textbf{Data Availability Statement} The data used to support the findings of this study are available from the corresponding author upon request.

\textbf{Conflict of interest} The authors declare that there is no conflict of interests regarding the publication of this paper.

\bibliography{iclr2021_conference}
\bibliographystyle{iclr2021_conference}

\end{document}

%% file: math_commands.tex
\usepackage{amsmath,amsfonts,bm}

\def\eqref#1{equation~\ref{#1}}
\def\1{\bm{1}}

\DeclareMathAlphabet{\mathsfit}{\encodingdefault}{\sfdefault}{m}{sl}
\SetMathAlphabet{\mathsfit}{bold}{\encodingdefault}{\sfdefault}{bx}{n}